\documentclass[letterpaper, 10 pt, conference]{ieeeconf}  %

\IEEEoverridecommandlockouts
\usepackage{times} %
\usepackage{amsmath} %
\usepackage{amssymb}  %
\usepackage{cite}
\usepackage{footnote}
\usepackage{comment}
\makesavenoteenv{tabular}
\usepackage[table,xcdraw]{xcolor}

\usepackage{multirow}
\usepackage{array}
\usepackage[T1]{fontenc}
\usepackage[utf8]{inputenc}
\usepackage{tabularx,ragged2e,booktabs}
\newcolumntype{C}[1]{>{\Centering}m{#1}}

\usepackage{makecell}
\usepackage[table,xcdraw]{xcolor}

\usepackage{nomencl}
\usepackage{relsize}
\usepackage{pdfpages}
\usepackage{caption}

\usepackage{color}
\usepackage{color, soul}
\usepackage{gensymb}
\usepackage{blindtext}
\usepackage{textcomp}
\usepackage{lipsum}

\makeatletter
\let\NAT@parse\undefined
\makeatother
\usepackage[hidelinks]{hyperref}
\usepackage{color}
\usepackage{wasysym}
\usepackage{subfig}
\usepackage{graphicx}
\usepackage{soul}
\usepackage{float}
\usepackage{amssymb}

\title{\LARGE \bf
An Autonomous, 3D Printed, Waterjet-Powered, Open-Source\\ Robotic Trimaran for Environmental Inspection and Monitoring}

\author{Reuben O'Brien, Martin Lambrechtse-Reid, Minas Liarokapis %
\thanks{Reuben O'Brien and Minas Liarokapis are with the New Dexterity Research Group, Department of Mechanical and Mechatronics Engineering, The University of Auckland, New Zealand.}
\thanks{Martin Lambrechtse-Reid is with the Department of Mechanical and Mechatronics Engineering, The University of Auckland.}
\thanks{E-mails: robr763@aucklanduni.ac.nz, martin.lambrechtse-reid@auckland.ac.nz, minas.liarokapis@auckland.ac.nz}%
}

\usepackage{eso-pic}
\begin{document}
\AddToShipoutPictureBG*{\AtPageLowerLeft{\makebox[\paperwidth][c]{\raisebox{0.22in}{\parbox[b]{\textwidth}{\scriptsize\centering \copyright~2024 IEEE. Personal use of this material is permitted. Permission from IEEE must be obtained for all other uses, in any current or future media, including reprinting/republishing this material for advertising or promotional purposes, creating new collective works, for resale or redistribution to servers or lists, or reuse of any copyrighted component of this work in other works. Accepted version. DOI: 10.1109/IROS58592.2024.10801622}}}}}

\maketitle

\begin{abstract}
Versatile, autonomous robotic boats can offer excellent environmental inspection and monitoring solutions for remote, dangerous, hard to reach, or access protected water bodies.
This paper introduces such a platform in the form of an autonomous, cost-effective, waterjet-powered robotic trimaran. Motivated by the need for an efficient aquatic monitoring, particularly in Aotearoa - New Zealand's diverse environments, the trimaran provides an efficient, low-cost, and easy to replicate alternative to resource-intensive research vessels. The proposed platform, costs \$600-1,500 USD to develop (depending on the sensing system configuration), weighs under 5 kg, and excels in bathymetry and water quality testing. The trimaran can reach speeds of up to 2 m/s offering obstacle avoidance of natural features, such as rocks. Utilizing off-the-shelf components and 3D printing technology, the proposed platform offers excellent reproducibility and robustness  while operating in shallow waters with its jet propulsion system. The paper presents in detail the design characteristics, the sensing system employed, testing results focusing on bathymetry, and highlights the ability of vessel and the potential for future research and data collection.
\end{abstract}
\section{Introduction}\label{sec:introduction}

Unmanned surface vehicles (USVs) have witnessed a surge in popularity in recent years, driven by advancements in autonomous technologies and their diverse applications in aquatic environments. These vehicles offer a promising avenue for scientific exploration, environmental monitoring, and resource management. The flexibility and adaptability of USVs make them particularly attractive for tasks ranging from data collection to environmental surveillance.

In the case of Aotearoa - New Zealand, the rivers and water streams constitute a significant 425,000 kilometres of length \cite{environmentguide-freshwater}, accompanied by a coastline length of 15,000 kilometers \cite{bell1996public}. This diverse aquatic landscape encompasses a myriad of ecosystems, each demanding monitoring to ensure longevity and health. In the Aotearoa - New Zealand Biodiversity Strategy\cite{biodivstrategy2020}, it details the key themes, with one being ecosystem protection, necessitating the application of seafloor and freshwater photogrammetry. This approach facilitates habitat mapping, characterization, and the study of aquatic species distributions. Presently, research in this field relies on manned research vessels, which are resource-intensive and incapable of accessing numerous freshwater rivers, streams, and wetlands.

\begin{figure}[ht]
    \centering
    \includegraphics[width=\columnwidth]{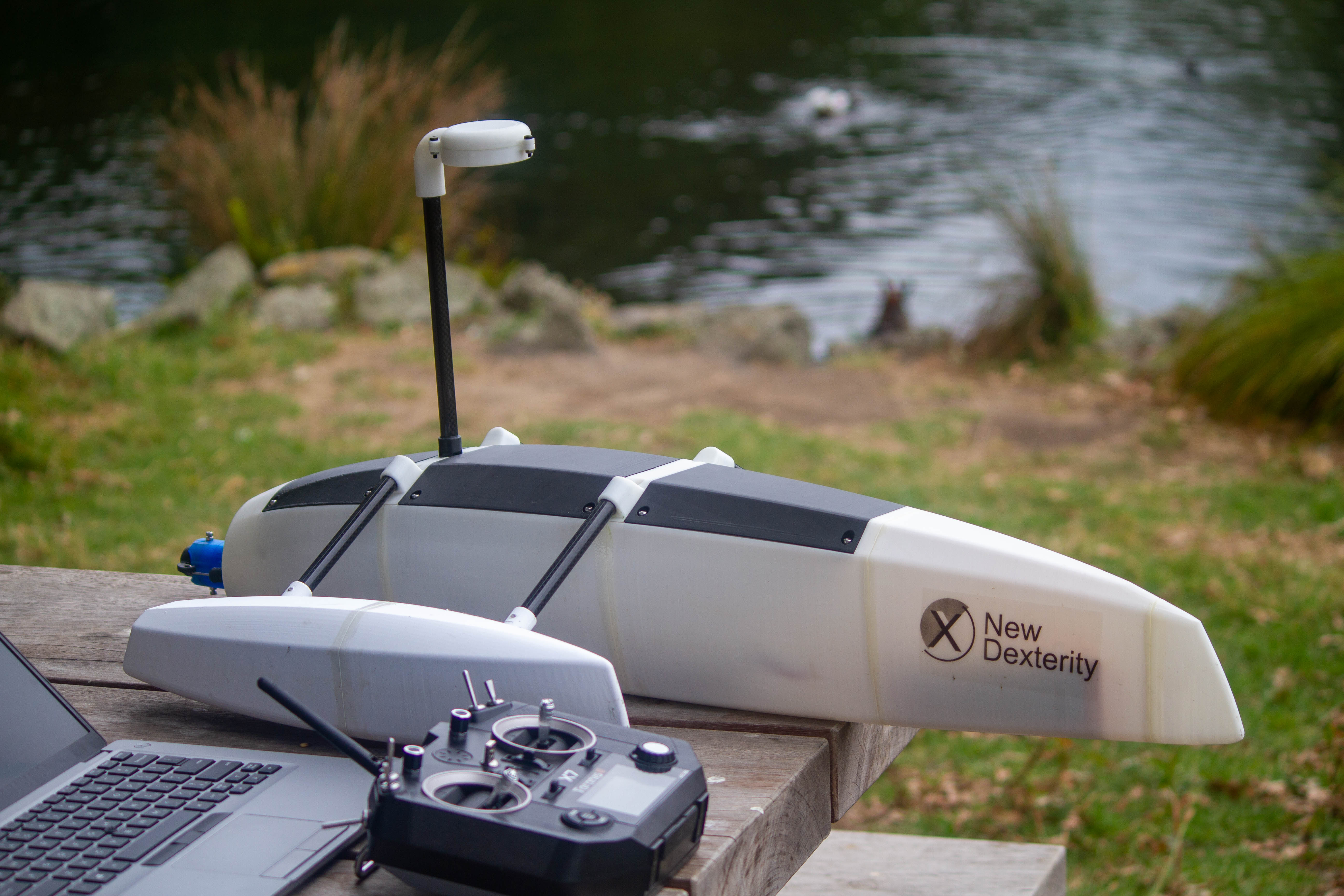}
    \caption{The proposed autonomous, 3D printed, waterjet-powered, open-source robotic trimaran.}
    \label{cover}
\end{figure}

To address the inefficiency of conventional systems, alternate solutions, notably USVs, have been proposed. The inaugural USV, developed in 1993 by MIT Sea Grant College Program \cite{Oceans}, marked the initiation of a field that has since witnessed the design of over 60 prototypes \cite{Schiaretti2017}. They each boast varying levels of autonomy and aquatic monitoring capabilities \cite{Schiarettii2017}. This count surpasses 60 when including vessels developed for USV competitions like RoboBoat \cite{August2020} and Robot X \cite{Park2017}, platforms where students actively contribute to advancing research and development in the field.

\begin{figure*}[!ht]
    \centering
    \includegraphics[width=1\textwidth]{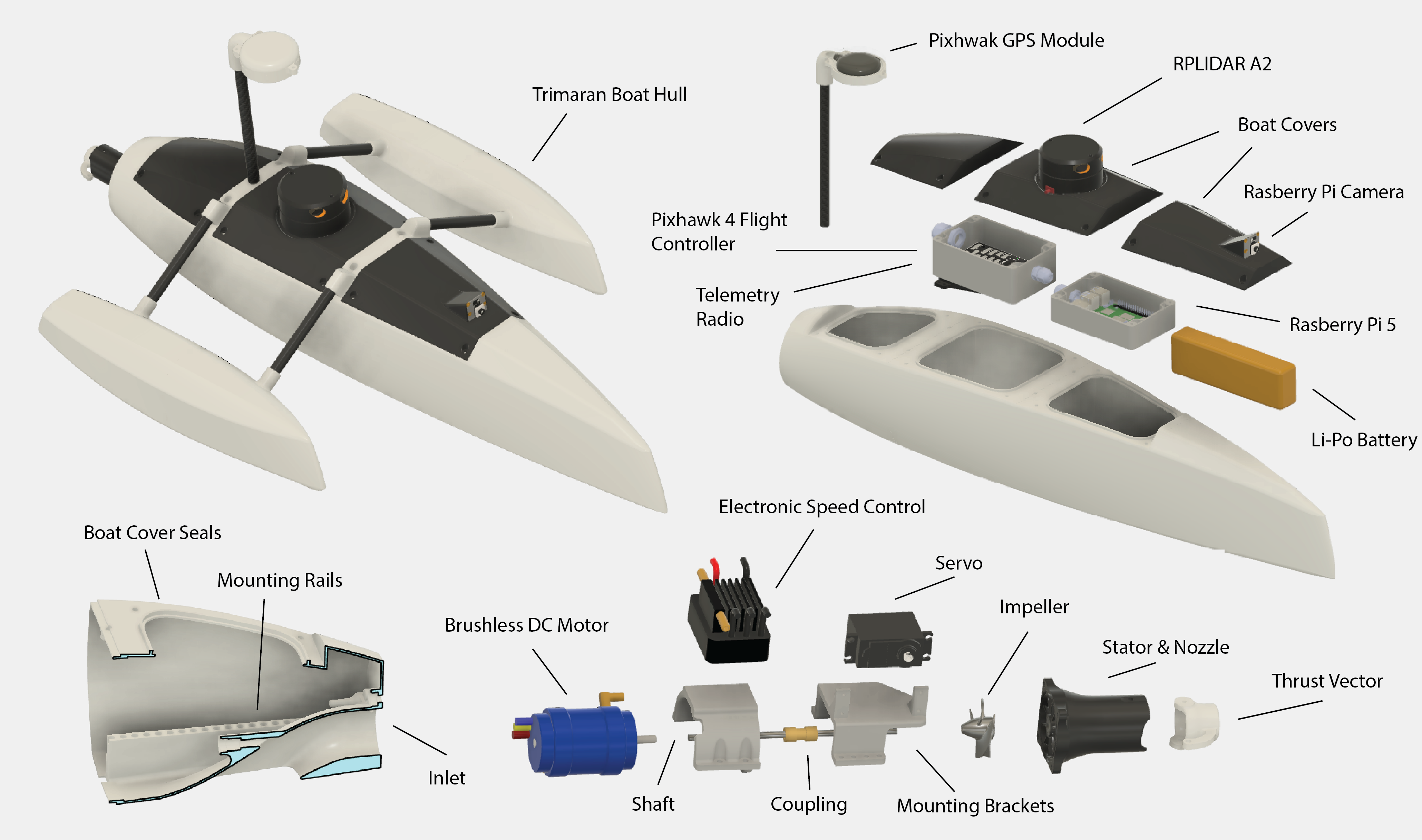}
    \caption{Exploded view of the proposed autonomous, 3D-printed, waterjet-powered, open-source trimaran platform.}
    \label{exploded}
\end{figure*}

Among the systems focusing on research, a significant number are open-sourced, with the majority featuring a fiberglass or plastic mouldered catamaran hull design \cite{Zhou2020, Carlson2023, Sornek2022, Carlson2019, Azevedo2016, Martinez2023}. These vessels are designed to be operated by 1-2 individuals and incorporate waypoint tracking capabilities. The overall cost of these USVs varies, starting from \$3,000 USD and increasing depending on the selected sensors \cite{Lee2023}. Notably, the most cost-effective and straightforward design utilizes an off-the-shelf plastic container as the hull, costing \$1,557 \cite{Ryu2022}. Commercial alternatives, such as the Blue boat USV \cite{bluerobotics2023}, are available but surpass \$4,200 USD, excluding training and ongoing maintenance costs.

For all USVs, effective trajectory mapping is crucial \cite{Chen2020}, as is the integration of sensors to prevent collisions \cite{Song2018}. Several papers have investigated strategies for interpreting data and determining optimal avoidance maneuvers when encountering other ships or objects \cite{Zhang2018, Chen2018}. Probabilistic approaches have been explored, relying on a framework to assess the mission capability of USV systems \cite{Zhang2023}. Common strategies encompass real-time analysis using LiDAR \cite{Lee2023} and cameras \cite{Sitorus}, employing algorithms such as YOLO \cite{Liu2021}. Moreover, the integration of swarm robotics \cite{GREGORY2020e0010}, \cite{9309408} further amplifies the capabilities of USVs, allowing for collaborative and coordinated efforts in data gathering and monitoring. This collective approach enhances the efficiency and coverage of surveillance operations in large water bodies, contributing to a more comprehensive assessment and understanding of complex aquatic ecosystems.

In response to the need for an affordable, easy to replicate, and robust platform, this paper proposes a USV powered by a waterjet system that has a development cost of \$600-1500 USD and that leverages off-the-shelf components and 3D printing for its manufacturing. The remainder of this paper is structured as follows: Section II sections delves into the design of the platform, Section III discusses the experimental validation and presents the results, while Section IV concludes the paper, discussing the proposed solution's potential.

\section{Design}\label{sec:design}

\begin{figure*}[ht]
    \centering
    \includegraphics[width=2\columnwidth]{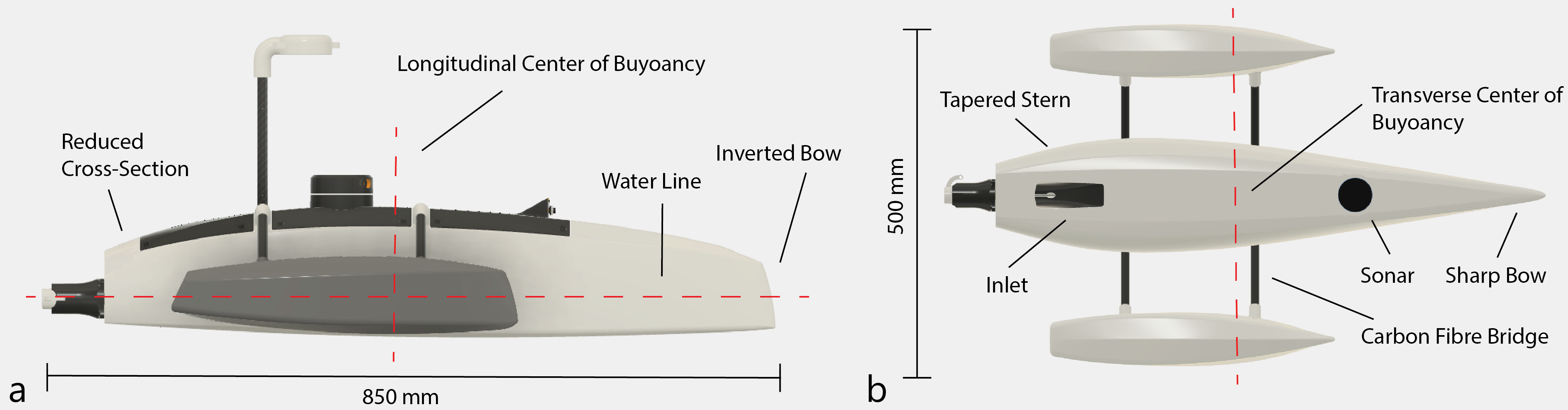}
    \caption{The side and top views of the proposed trimaran platform. Subfigure (a), presents a side view highlighting the inverted bow. Subfigure (b), presents the bottom of the hull, displaying the tapered stern and jet inlet.}
    \label{hull}
\end{figure*}

The proposed trimaran system was design to demonstrate effective navigation in freshwater. The budget is limited to \$1,500 USD, requiring the use of cost-effective consumer-grade technologies. Regarding design specifications, the trimaran's size was set to not exceed 1 m in length, 0.5 m in width, and 5 kg in weight for single-person handling. It was determined that the platform should operate at an average speed of 1.5-2 m/s with the ability to go faster if needed. These requirements guide the design, development, and testing phases of the project.

\begin{table}[h]
\caption{Trimaran Components}
\label{table1}
\centering
\begin{tabular}{lccc}
\rowcolor[HTML]{BFBFBF}
\multicolumn{1}{c}{\cellcolor[HTML]{C0C0C0}{\color[HTML]{000000} Part}} & \multicolumn{1}{l}{\cellcolor[HTML]{BFBFBF}Qty} & \multicolumn{1}{l}{\cellcolor[HTML]{BFBFBF}Cost (USD)} & \multicolumn{1}{l}{\cellcolor[HTML]{BFBFBF}Weight (kg)} \\
\rowcolor[HTML]{F2F2F2}
\multicolumn{4}{c}{\cellcolor[HTML]{F2F2F2}Powertrain}           \\
BLDC motor 3660 3180\,kv & 1 & \$98.69 & 0.24 \\
ESC 120\,A & 1 & \$105.79 & 0.15 \\
LiPo 3S 80C 5000\,mAh & 1  & \$175.21 & 0.46 \\
LiPo 6S 80C 5000\,mAh & 1  & \$75.00 & 0.74 \\
Drive shaft coupling & 1 & \$5.56 & - \\
Flange ball bearing  & 2 & \$7.10 & - \\
CNC aluminium impeller   & 1 & \$21.60 & - \\
Servo stall torque  3\,kg & 1 & \$13.10 & 0.04 \\
Power Module (12S) & 2 & \$37.98 & 0.04 \\
\rowcolor[HTML]{F2F2F2}
\multicolumn{4}{c}{\cellcolor[HTML]{F2F2F2}Electronics}\\
Polycarbonate Enclosure IP67 & 2 & \$90.00 & 0.30 \\
Rasberry Pi 5 & 1 & \$100.28 & 0.05 \\
Pixhawk 4 mini + GPS module & 1 & \$201.00 & 0.03 \\
Telemetry Radio & 1  & \$40.61 & 0.03 \\
RPLIDAR A2 & 1 & \$319.00 & 0.22 \\
Ping2 Sonar & 1 & \$435.00 & 0.19 \\
Rasberry Pi camera v3 & 1 & \$41.78 & - \\
\rowcolor[HTML]{F2F2F2}
\multicolumn{4}{c}{\cellcolor[HTML]{F2F2F2}Miscellaneous}\\
Carbon Tube (1m) & 1 & \$11.36 & 0.03 \\
PETG 3D printing filament (1\,kg) & 2 & \$40.00 & 1.25 \\
Cable gland & 3 & \$9.24 & - \\
Threaded inserts (100) & 1 & \$21.47 & - \\
\rowcolor[HTML]{F2F2F2}
\multicolumn{4}{c}{\cellcolor[HTML]{F2F2F2}Cost: \$1555.28, Weight: 3.77kg}
\end{tabular}
\end{table}

The maximum development cost of the boat is \$1,533.68 USD (as seen in Table \ref{table1}) with the minimum viable cost being \$592.62 USD, if all the non-crucial sensors are removed. These extras components are the Raspberry Pi, camera, LiDAR, and the Sonar, resulting in an autonomous vessel that relies purely on GPS and IMU positioning.

\begin{figure}[ht!]
    \centering
    \includegraphics[width=\columnwidth]{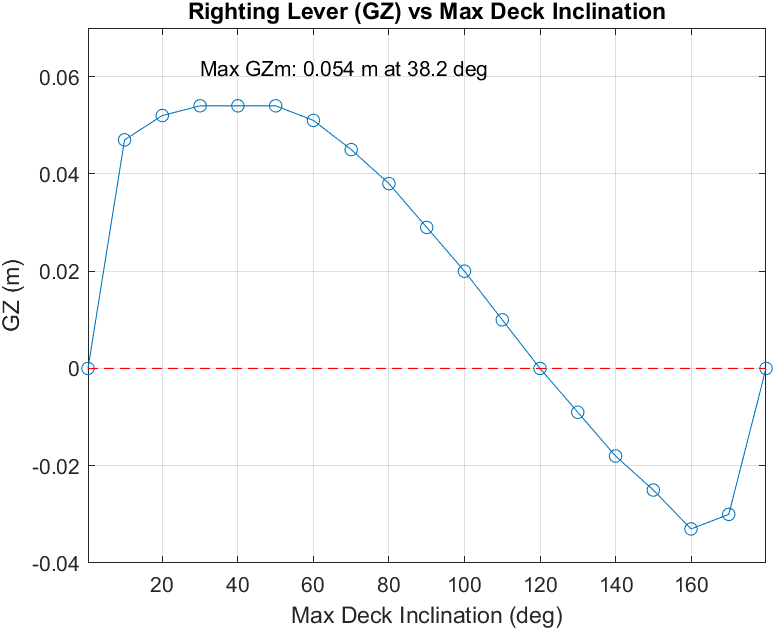}
    \caption{Righting lever of the trimaran hull is seen at a range of deck inclination levels (roll angles). This highlights that the peak lever is at 38.2 deg and the point of negative lever (unrecoverable roll angle) is at 120 deg.}
    \label{righting}
\end{figure}

\subsection{Trimaran Design}
Considering the aforementioned specifications, opting for a mono-hull design would introduce inherent instability in the roll axis, hindering the effective operation of sensing electronics. While catamaran hulls offer better roll stability, they often require dual propulsion systems, contributing unnecessary weight and power consumption for the imposed specifications. Consequently, a trimaran configuration (see Fig. \ref{exploded}) was chosen for its roll stability akin to a catamaran but with the advantage of requiring a single propulsion system. The proposed trimaran has been optimized to dimensions of 850 mm in length and 500 mm in width, tailored for a weight of 5 kg and speeds of 2 m/s. The detachable side hulls are affixed using bolts and enable convenient transport, resulting in a compact package size of 850 mm x 250 mm x 200 mm.

\begin{figure}[ht!]
    \centering
    \includegraphics[width=0.95\columnwidth]{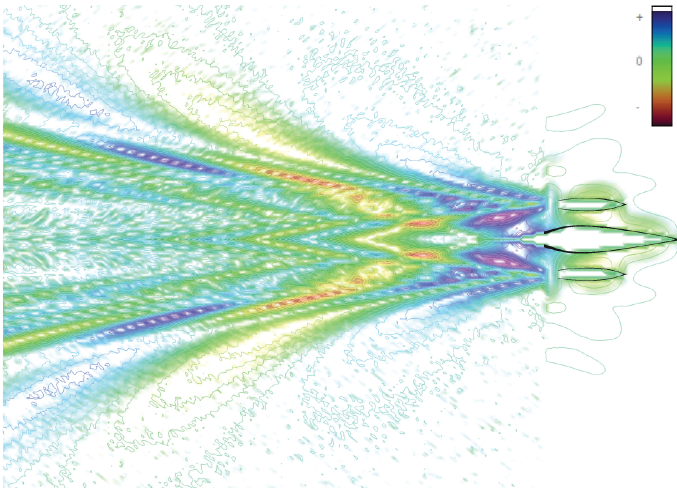}
    \caption{Free surface wave pattern generated by the trimaran at a speed of 2 m/s, ignoring the affects of viscosity and wave breaking (idealised calculation). The simulations shows how the waves from each hull interact, with a peak of 50 mm and a trough of 40 mm from the water line.}
    \label{wake}
\end{figure}

\subsubsection{Trimaran Form}

Figure \ref{hull}a illustrates the gradual reduction in cross-sectional area at the stern, while Figure \ref{hull}b depicts the tapering from the underside. This design is crucial for a displacement hull that remains submerged throughout all operational speeds. The tapered hull enhances efficiency by ensuring a smooth flow of water, minimizing the creation of vortices at the ends. The inverted bow in figure \ref{hull}a, is similar to that of combatant naval hull forms \cite{10.5957/WMTC-2015-038}, such as the US Navy destroyer Zumwalt 1000, along with the Mayflower, a autonomous research vessel \cite{rivkin2021unmanned}. This design choice aims to reduce the wake around the boat and prevent pitching by piercing the waves rather than riding up and over them \cite{khoob2023experimental}. While alternative designs, such as a axe bow are shown to be effective, they were not selected due to the anticipated pitch of the hull caused by the propulsion system. This alternative nose design would lead to increased water riding up the sides of the boat. However, being unmanned, this area is completely sealed off with a constant curve to facilitate water runoff rather than accumulation.

To minimize the submerged hull height, a flat bottom is used, tapering up from the edges of the jet inlet. This design increases buoyancy at shallower depths while providing a smooth surface to prevent snagging and local damage, which could be prevalent with the use of a keel or a sharp V bottom.

\begin{figure*}[ht]
    \centering
    \includegraphics[width=2\columnwidth]{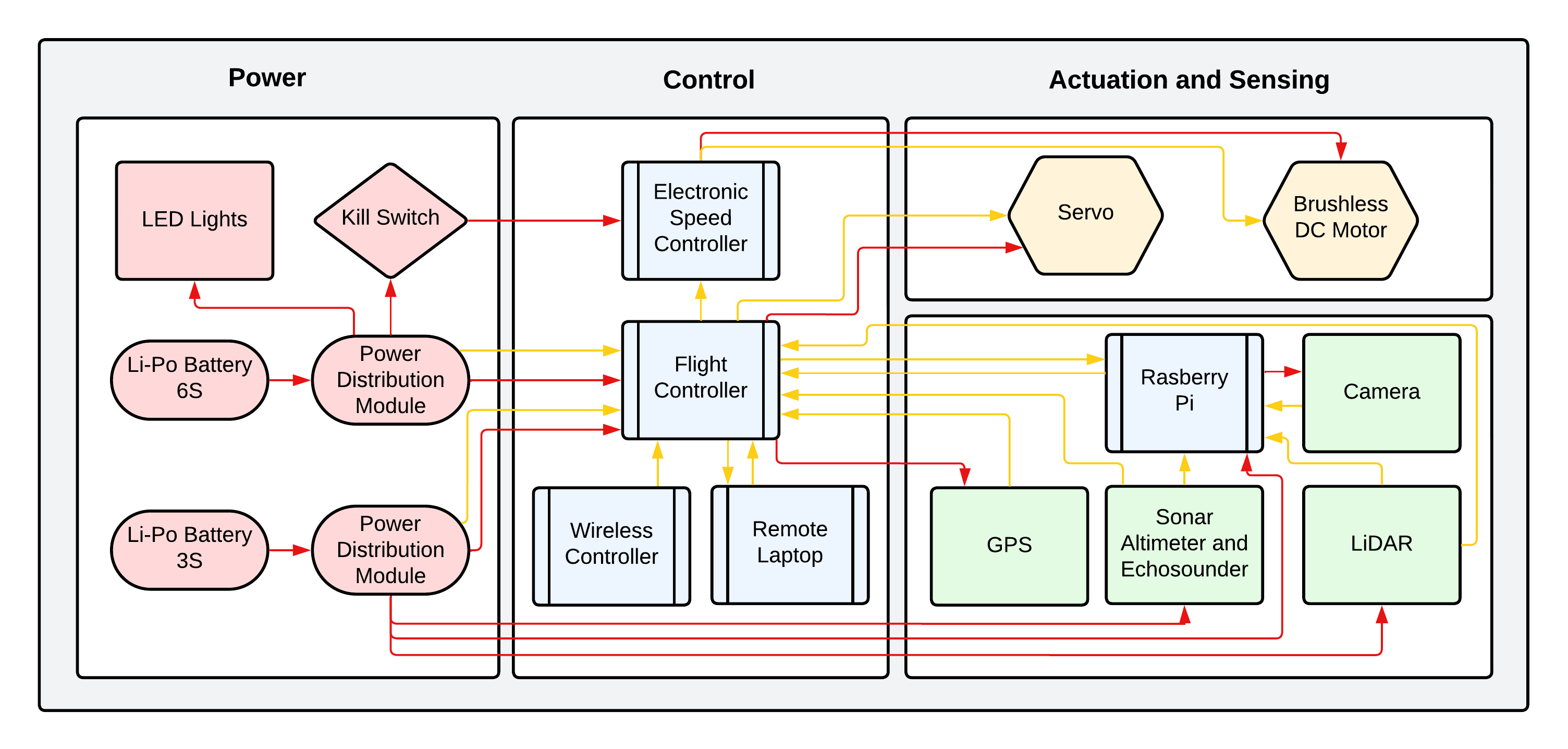}
    \caption{A block diagram of all the key electronics that make up the proposed autonomous trimaran platform. There are three main categories: i) Power (red), containing batteries and power modules, ii) Control (blue), representing the core component: the flight controller. Lastly, iii) Actuation (yellow) and Sensing (green), representing the motors along with the vision and algorithmic processing components.}
    \label{flow}
\end{figure*}

\subsubsection{Simulation}

Simulation analysis of the trimaran hull was done in the MAXSURF software that allowed for mesh creation, followed by hydro-static and resistance analysis. In Figure \ref{hull} a side on view of the hull can be seen which outlines the waterline of the hull at 46 mm which is achieved at an overall weight of 5 kg. If the weight of the hull is reduced the waterline lowers along with the center of buoyancy, therefore the hull will ideally always operate at this weight. To ensure the center of mass is lower than the center of buoyancy the electronics are layed out along the length of the hull, along the base. They are positioned according to the longitudinally center of buoyancy (LCB) which is located at 310 mm from the stern. The LCB is the point at which buoyancy acts upwards against the hull. The transverse center of buoyancy is located along the center-line due to the symmetry of the hull. When electronics are mounted inside the hull of the platform, it is crucial that the center of gravity lines up with this in both directions so as to ensure a stable design.

The righting level \ref{righting} calculation was performed with the hull at a weight of 5 kg at a height of 30 mm from the base which can be achieved by mounting the batteries to the base. If the weight is reduced the peak lever will decrease from the current peak at 38.2 deg. According to \cite{wartsila-intact-stability} a righting lever shall be more than 30 degrees but no less than 25 degrees. The point of vanishing stability is at 120 degrees of inclination, which will allow this hull to perform well in all weather conditions.
Reducing the wake of a hull, reduces the resistance acting on the hull allowing for greater speeds and increased efficiency. Analysing the predicted wake, \ref{wake} the greatest disturbance occurs directly behind the hull in three separate forms from each hull.

As the disturbance moves back the waves build upon one another creating peaks and troughs to the edges. More than a meter from the stern the wake disperses from the center and leaves it relatively undisturbed.

\subsubsection{Propulsion}
Safety is the major concern here, with the proposed robotic vessel being unmanned and the goal of conducting unsupervised inspection and monitoring operations of unstructured environments. Since the size of the proposed platform is relatively small, people may try to interact with it and the use of a regular propeller driven system could lead to injuries. This is not the case with a jet impeller that is completely housed. Since it is housed stronger materials can be used such as stainless steel for the impeller which allows for improved durability and efficiency compared to plastic alternatives. Jet drive systems also allow for no additional clearance under the submerged hull, allowing operation in depths of 60 mm onwards.

Turn radius compared to alternate systems is greatly improved from similar sized systems from 2.5 m \cite{Martinez2023} and 5 m \cite{Carlson2019} to 1 .5 m. When the jet is under power the thrust is all directed in the way of the single vectoring nozzle. This is a significant improvement over prop wash gained from fixed propeller using rudders, and results in similar performance to skid steering from the use of multiple propulsion systems. Skid steering was not selected due to the requirement of two independent motors and ESCs. One of the downfalls with jet drives is their inability to turn when no thrust is applied. Another is reversing which can only be achieved by using an actuated bucket over the nozzle \cite{Mitchell2022} that redirects the jet stream back under the boat. This allows for speeds up to 0.5 m/s but it is an inefficient and unstable system.

\begin{figure*}[ht]
    \centering
    \includegraphics[width=1.93\columnwidth]{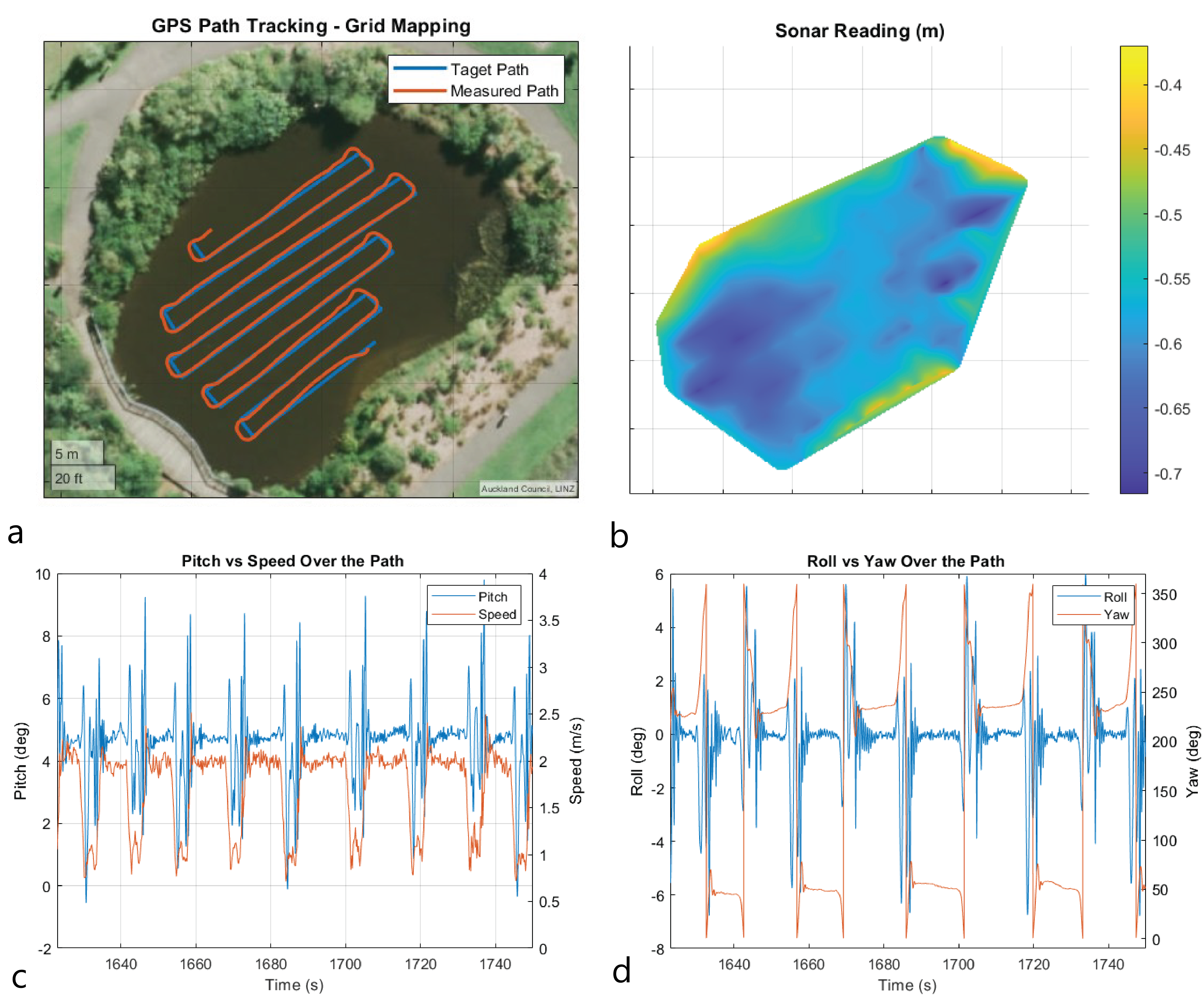}
    \caption{GPS tracking of the autonomous trimaran at the Newmarket lake in Auckland, New Zealand (36°51'52.2"S, 174°47'00"E) is shown on subfigure (a) with the measured path against the target path. This route contained 9 straights of 15 to 20 m with nine 2 m long joining sections. Subfigure (b) details the sonar data collected over the path showing a depth range of 0.3-0.7 m. The data displayed in subfigures (c,d) are from the route traverse. Subfigure (c) is detailing the pitch range of the hull (0 to 9 degrees), with a steady 4 degrees as it maintained the target speed of 2 m/s. Subfigure (d) highlights the minimal roll (-6 to 6 degrees) of the hull as it performs 90 degree turns between the straights. }
    \label{experiment}
\end{figure*}

A brushless DC motor was chosen with the key considerations being the expected power output and speed with a jet pump where available torque is more important than the maximum rotational speed. Therefore a lower KV motor is preferred in the range of 1000-4000 kv. To reduce current draw to the motor and ESC while still having the required power ouptut, a 6s battery was selected. Cooling of the motor and ESC is achieved through water cooling that is connected to the inlet of the jet and exits besides the jet stream.

\subsubsection{Fabrication}

The trimaran design is optimized for 3D printing on consumer-grade printers with a minimum print bed size of 250 mm x 210 mm x 210 mm. Consequently, all individual parts, are tailored to fit within these dimensions. While this choice constrains the overall hull size of the proposed trimaran platform, a multi-part design ensures the required payload size is still achievable.

Material selection for 3D printing includes PLA, ABS, PETG, and TPU. PLA, while easy to use, lacks UV resistance and is prone to degradation and warping in outdoor conditions, particularly in direct sunlight, making it unsuitable for this application. ABS addresses these issues but introduces complexity and the need for add-ons like an enclosure during printing. Therefore, PETG emerges as the preferred material, overcoming PLA's drawbacks without the added challenges of printing ABS.

Although TPU is not suitable for the main hull due to a thin wall thickness of 1.5 mm, it is ideal for the nose piece, being capable of deforming and absorbing energy upon impact. Sections are securely attached using epoxy applied to overlapping flanges detailed into each part. Additional components, including covers, electronic brackets, and side hulls, utilize heated inserts for easy removal using M3 bolts.

\begin{figure}[ht!]
    \centering
    \includegraphics[width=\columnwidth]{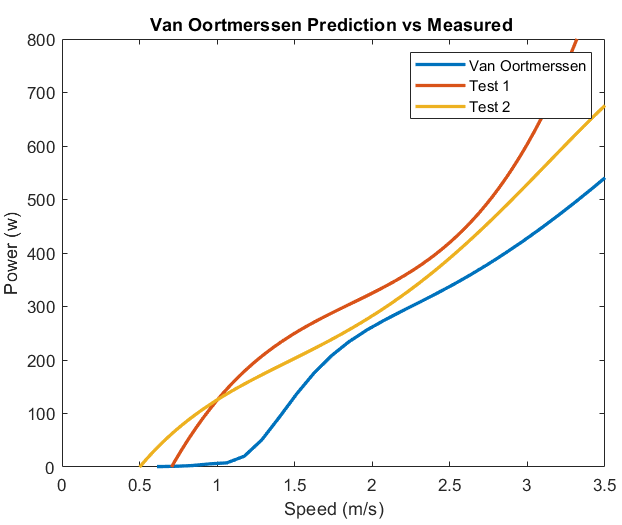}
    \caption{A comparison between the predicted power required at varying speeds against the measured power during testing.}
    \label{van}
\end{figure}

Alternative methods of manufacture include 3D printing negatives of the hull to be able to create fiberglass bodies that are significantly more robust and result in a lower weight. This can lead to an increased ability to carry more sensing or measuring electronics. The caveat is that the skills, materials, and equipment required to create a fiberglass hull are specialized, therefore, this method was excluded. Hulls such as \cite{Carlson2023} vacuum formed around a mould creating a single piece for the submerged region of the hull. This requires access to a vacuum former of over 1 m$^2$.

\subsection{Control and Sensing Hardware}

To power the trimaran there are two independent power sources which allow the flight controller and the GPS to have redundant power as seen in figure \ref{flow}. Therefore if the auxiliary (LIPO 3S) battery dies the system can still return in either the auto mode through GPS navigation or on manual through the controller. If the power supply for propulsion (6s) runs out, it is still capable of locating itself and sending these commands to a the remote data collection laptop.

\begin{figure}[ht!]
    \centering
    \includegraphics[width=\columnwidth]{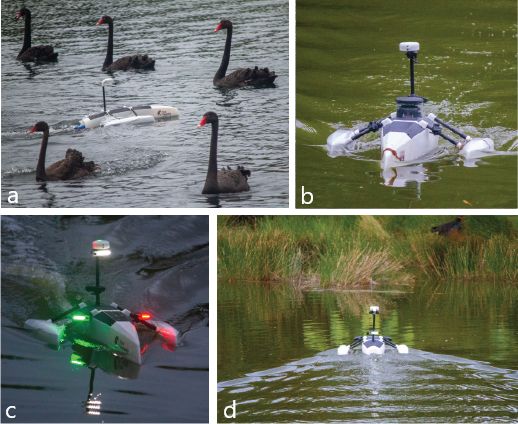}
    \caption{Autonomous operation experiments. Subfigure (a) presents the trimaran operating amongst black swans, showcasing the small scale of the vessel. Subfigure (b), demonstrates that the bio-fouling present during testing, was of no hindrance. Subfigure (c), shows that the platform can operate efficiently in low-light and at night time conditions thanks to the available lighting that allows other vessels to interpret the direction of travel and prevent a collision. Lastly, subfigure (d) shows the wake created from the hull which closely resembles the estimated wake during the simulation.}
    \label{test}
\end{figure}

The flight controller running Ardupilot is the central unit and all the commands are going through it to the propulsion system. Key inputs are the wireless controller that is capable of switching the mode of the system from auto to manual so as to allow for recovery in case of autonomous control failure. While in auto mode the GPS and built-in IMU in the flight controller allow it to locate and orientate itself within 2 m of accuracy in outdoor environments. Raw data from the LiDAR is processed within the flight controller and is filtered into 4 segments that provide a 180° field of view to prevent obstacle collisions.

The Raspberry Pi acts as the data logger for the camera and any other sensors on-board such as an sonar echo-sounder.
To control the trimaran a proportional and integral (PI) controller is used to control the steer rate and thrust response relative to the desired location and speed set point. The flight controller is capable of doing this, but to use more complex controllers such as model predictive controllers \cite{Lee2023} the Rasberry Pi is needed.

\section{Experiments}\label{sec:results}

\subsection{Power and Speed}
As seen in Figure \ref{van} the power used throughout testing with the proposed trimaran platform was plotted against speed and compared to the prediction model generated within MAXSURF. The employed Van Oortmerssen model was designed for small, high speed vessels designed to operate in displacement \cite{Oosanaen}. The predicted results are shown in Figure \ref{van}, and assume an overall propulsion efficiency of 40\% which is expected \cite{Mitchell2022}. Test 1 showed a similar output in regard to the gradient, but was 17\% off at 2 m/s. Test 2 at the desired speed of 2 m/s, the predicted is 8\% less than that measured, with the gradient diverging before and after this. This makes it a representative model within 20\% for future speeds and designs of scaled models.

Therefore, at 1 m/s there will be a constant draw of 125 W allowing for 1 hr of continuous run time, with this reducing to 38 minutes at 1.5 m/s and 27 minutes at 2 m/s with the current 6S 5000 mAh LI-PO battery. The trimaran system was designed for 5 kg and only weighed 3.8 kg during testing, therefore a 10,000 mAh could be used. This adds an additional 0.7 kg (4.4 kg total) and would double the available run time at each speed.

\begin{figure}[ht!]
    \centering
    \includegraphics[width=\columnwidth]{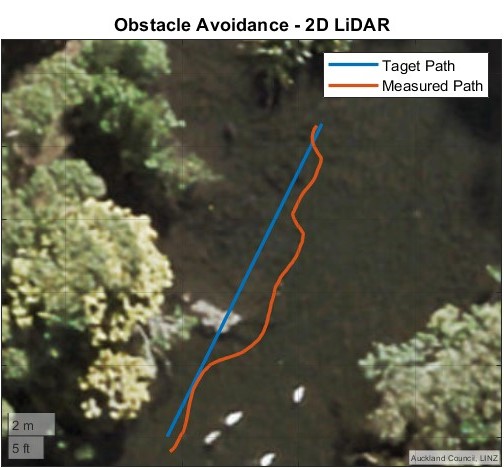}
    \caption{Obstacle avoidance of rocks located in the Western Springs lake in Auckland, New Zealand (36°51'55.7"S, 174°43'19.6"E) using the 'bendy-ruler' algorithm from the 2D LiDAR data.}
    \label{avoid}
\end{figure}

\subsection{Platform Testing}
During the experiments shown in Fig. \ref{test} the jet was capable of running without being entangled by bio-fouling in the middle of the pond. Around the edges the algae and plants density increased, resulting in plants being wrapped around the input shaft but the jet was always capable of running. This was thanks to the CNC aluminium machined impeller that was able to free itself from anything that was caught in it. However, branches and alagae got caught around the sharp fronts of the hull and around the jet housing. Their effect was minimal but could lead to reduced efficiency over extended run time. At speeds over 2 m/s water was washing riding up and over the side hulls leading to spray coming from the mounting tubes, but the bow stayed under the surface of the water as intended. The minimal wake generated due to the inverted bow and tapered stern created little disturbance to the water allowing testing around marine life.

\subsection{Autonomous Testing}

The satellite image in Figure \ref{test}a is out of date and doesn't show the vegetation overgrowth that is now present. It appears that the boat should be able to traverse more of the pond, however, the target path is to the edge. This highlights the value in the combination of both local and global path planning. If the path was planned remotely it would of been impossible to complete without the use of local obstacle avoidance to stay clear of the overgrowth. Analysing the path tracking of the trimaran through the grid mapping in figure \ref{experiment}a it is possible to complete the path with the largest error recorded being 1.6 m. The error is seen during left hand turns where it appears to be unable to turn as tight as in the case of the right hand turns. Figure \ref{experiment}d confirms this with the hull rolling slightly more to the right creating an asymmetric performance. This may be due to weight distribution within the hull or the trim adjustment of the vectoring nozzle. However, in both directions there is very little roll, with a total range of 12 degrees. During periods of constant yaw the hull displays zero roll.

The hull is capable of tracking the 2 m/s during the straights with fluctuation between 1 m/s and 2.5 m/s during and directly after the corners. At a constant speed there is a 4 degree pitch as seen in Figure \ref{experiment}c with it raising and lowering within a 9 degree range, peaking at 9 degrees. Pitch response of the hull is directly proportional to the speed, with a deceleration resulting in a reduction and an acceleration causing the nose to pitch up. Both the trimaran hull roll and pitch demonstrate the ability to capture accurate 2D LiDAR data that will be usable for navigating and avoiding local obstacles (rocks) in calm water conditions.

In \cite{Wang2020}, a complex system that is capable of performing SLAM, with the use of a 3D LiDAR through a man-made canal was proposed. A simplistic version of this is shown in Figure \ref{avoid} where the trimaran is able to traverse around rocks in a natural pond. This is done dynamically through the 2D LiDAR data where a 'bendy-ruler' algorithm is used to find a path that can be traversed without obstacles. The look ahead distance was set to 5 m with the avoidance distance set to 2 m. The LiDAR is capable of detecting objects up to 8 m afar indoor but outdoor in water it was reliable at distances up to 5 m. An alternate YDLiDAR G4 was tested as it had a greater range of 16 m, but the results were worse. It was unable to detect black surfaces outdoors at a distance of 5 m. Using a high performance laser with wavelengths greater than 785 nm would be beneficial in reducing interference with sunlight.

\subsection{Video, CAD Files, Code}

A video presenting the experimental validation of the performance of the trimaran as well as the CAD files and the code can be found at the following URL:
\begin{center}
\href{http://www.newdexterity.org/trimaran}{www.newdexterity.org/trimaran}
\end{center}
We use an open-source dissemination of the results to allow replication and further development by others.

\section{Conclusions}\label{sec:conclusions}

In conclusion, the presented autonomous robotic trimaran emerges as a cost-effective and innovative solution for aquatic inspection and monitoring, particularly in the diverse water bodies of Aotearoa - New Zealand. Through a comprehensive presentation of the design and build considerations and careful sensor selection, we have demonstrated the feasibility of creating an efficient, affordable, 3D printed, easy to replicate robotic platform. The successful bathymetry and water quality testing at speeds of 2 m/s underscore the practical utility of this technology. This research contributes to the field by providing a light-weight, low cost, open-source, waterjet-powered trimaran, offering a promising alternative to resource-intensive research vessels.

Looking forward, potential enhancements and refinements could include full integration and fusion between the LiDAR and camera to allow the platform to operate in un-mapped and overgrown environments. Moreover, deriving the dynamic model of the robotic boat so as to allow for it's use in more complex scenarios with more sophisticated algorithms aiding in path planning and obstacle avoidance, will also be a future direction.

\bibliographystyle{IEEEtran}
\bibliography{REF}
\end{document}